\pdfoutput=1

\documentclass[11pt]{article}
\usepackage[table]{xcolor}
\usepackage{acl}
\usepackage{multirow, makecell}
\usepackage{tabularx}
\usepackage{times}
\usepackage{latexsym}
\usepackage{booktabs,siunitx,multirow}
\usepackage{amsmath}
\usepackage[T1]{fontenc}
\usepackage{graphicx} 
\usepackage{subcaption}
\usepackage{booktabs}
\usepackage{todonotes}

\usepackage[utf8]{inputenc}
\usepackage{xspace}
\usepackage{microtype}
\usepackage[usestackEOL]{stackengine}
\newcommand{\methodone}{LRR\xspace}
\newcommand{\methodtwo}{NPR\xspace}
\newcommand{\methodonelong}{DetectLLM-LRR\xspace}
\newcommand{\methodtwolong}{DetectLLM-NPR\xspace}
\newcolumntype{P}[1]{>{\centering\arraybackslash}p{#1}}
\newcommand{\splitcell}[1]{\Centerstack{#1}}
\title{DetectLLM: Leveraging Log Rank Information\\ for Zero-Shot Detection of Machine-Generated Text}

\author{Jinyan Su$^1$,
  Terry Yue Zhuo$^2$,
  Di Wang$^3$,
  Preslav Nakov$^1$\\
  $^1$Mohamed bin Zayed University of Artificial Intelligence\\
$^2$Monash University and CSIRO's Data61,
  $^3$King Abdullah University of Science and Technology \\
  \texttt{\{Jinyan.Su, preslav.nakov\}@mbzuai.ac.ae}\\\texttt{terry.zhuo@monash.edu,di.wang@kaust.edu.sa}\\
  }

\begin{document}
\maketitle
\begin{abstract}
With the rapid progress of Large language models (LLMs) and the huge amount of text they generated, it becomes more and more impractical to manually distinguish whether a text is machine-generated. Given the growing use of LLMs in social media and education, it prompts us to develop methods to detect machine-generated text, preventing malicious usage such as plagiarism, misinformation, and propaganda. Previous work has studied several zero-shot methods, which require no training data. These methods achieve good performance, but there is still a lot of room for improvement.
In this paper, we introduce two novel zero-shot methods for detecting machine-generated text by leveraging the log rank information. One is called \methodonelong, which is fast and efficient, and the other is called \methodtwolong, which is more accurate, but slower due to the need for perturbations. Our experiments on three datasets and seven language models show that our proposed methods improve over the state of the art by 3.9 and 1.75 AUROC points absolute. Moreover, \methodtwolong needs fewer perturbations than previous work to achieve the same level of performance, which makes it more practical for real-world use.
We also investigate the efficiency--performance trade-off based on users preference on these two measures and we provide intuition for using them in practice effectively. We release the data and the code of both methods in \url{https://github.com/mbzuai-nlp/DetectLLM.}\end{abstract}

\section{Introduction}

Large language models (LLMs) have made rapid advancement in recent years, and are now able to generate text with significantly improved diversity, fluency, and quality. Models such as ChatGPT \cite{Chatgpt}, GPT-3 \cite{brown2020language}, LLaMa \cite{touvron2023llama} and BLOOM \cite{scao2022bloom} demonstrate exceptional performance in answering questions \cite{robinson2022leveraging}, writing stories \cite{fan2018hierarchical,yuan2022wordcraft}, composing emails, analyzing program code, and thus facilitating daily life and improving work efficiency. 
However, LLMs can also be misused for generating plagiarized text, misinformation, and propaganda, which  can lead to negative consequences.
For instance, 
students might use LLMs to write their essays and assignments \cite{rosenblatt2023chatgpt}, making fair evaluation difficult for teachers, and in the long run, undermining the integrity of the entire education system.  Malicious actors might generate fake news articles to spread misinformation and propaganda or to manipulate the public opinion, which is dangerous, especially when it comes to politics \cite{floridi2020gpt,stokel2022ai}. 

With the proliferation of LLMs and the increasing amount of texts it produced, it is challenging for humans to accurately identify machine-generated texts \cite{gehrmann2019gltr}. Moreover, it is unrealistic to hire humans to manually identify machine-generated text at scale due to the prohibitively high costs and the efficiency requirements in real-time applications, e.g.,~in social media.
Thus, it is essential to develop tools and strategies to automatically identify machine-generated text and to mitigate the potential negative impact of LLMs.

The problem of distinguishing machine-generated from human-written text is commonly formulated as a binary classification task \cite{jawahar2020automatic}. Most previous work has focused on the black-box scenario, where the detector has access to the output of the LLMs only and cannot make use of its internal representations and states. They train or fine-tune a supervised binary classification model using the output of the LLMs. 
Such methods lack flexibility since they need to be retrained from scratch to be able to recognize the output of a new LLM \cite{mitchell2023detectgpt}. Given the speed at which new LLMs are developed, black-box methods are becoming more and more expensive and impractical. In cases when the access to the LLM is via an API only, one possibility is for the LLM owner to record all content it has generated, or to watermark all texts it has generated \cite{kirchenbauer2023watermark, zhao2023protecting}.
However, such solutions are not feasible for third-parties.

We therefore consider a white-box setting, where the detector has full access to the LLMs. More specifically, we focus on zero-shot methods, where we use the LLM itself, without additional training. Generally speaking, zero-shot methods uses the source LLM to extract statistics such as the average per-token log probability or the average rank of each token in the ranked list of possible choices, and then to make a prediction by comparing it to a threshold \cite{solaiman2019release, ippolito2019automatic, gehrmann2019gltr}. Recently, \citet{mitchell2023detectgpt} observed that machine-generated text tends to lie in the negative curvature of the log likelihood of the text, proposing a perturbation-based zero-shot method called DetectGPT and achieving the best performance at the expense of efficiency.

Here, we introduce two novel zero-shot methods, which extensively exploit the potential of the log rank information. The first one is using Log-\textbf{L}ikelihood Log-\textbf{R}ank \textbf{r}atio (\methodone), which complements Log-Likelihood with Log-Rank to enhance the performance. The second one uses a \textbf{N}ormalized \textbf{p}erturbed log \textbf{r}ank (NPR), which is based on the intuition that machine-generated texts are more sensitive to minor rewrites (or say, small perturbations). We called these two methods \methodonelong and \methodtwolong respectively.

In summary, our contributions are as follows:
\begin{itemize}
\item We propose two novel zero-shot approaches based on log rank statistics, which improve over the state of the art. On average, the proposed two methods improved upon the previous best zero-shot methods by 3.9 and 1.75 AUROC points absolute.
\item We investigate the efficacy of existing zero-shot methods and explore their boundaries and limits as the size of the LLMs increases from 1.5 to 20 billion. 
\item We conduct comprehensive experiments to better understand the efficiency--performance trade-offs in zero-shot methods, thereby providing interesting insights on how to choose among different categories of zero-shot methods based on users' preference on performance or efficiency. 
\end{itemize}

\begin{figure}[t!]
    \centering
    \includegraphics[width=0.5\textwidth]{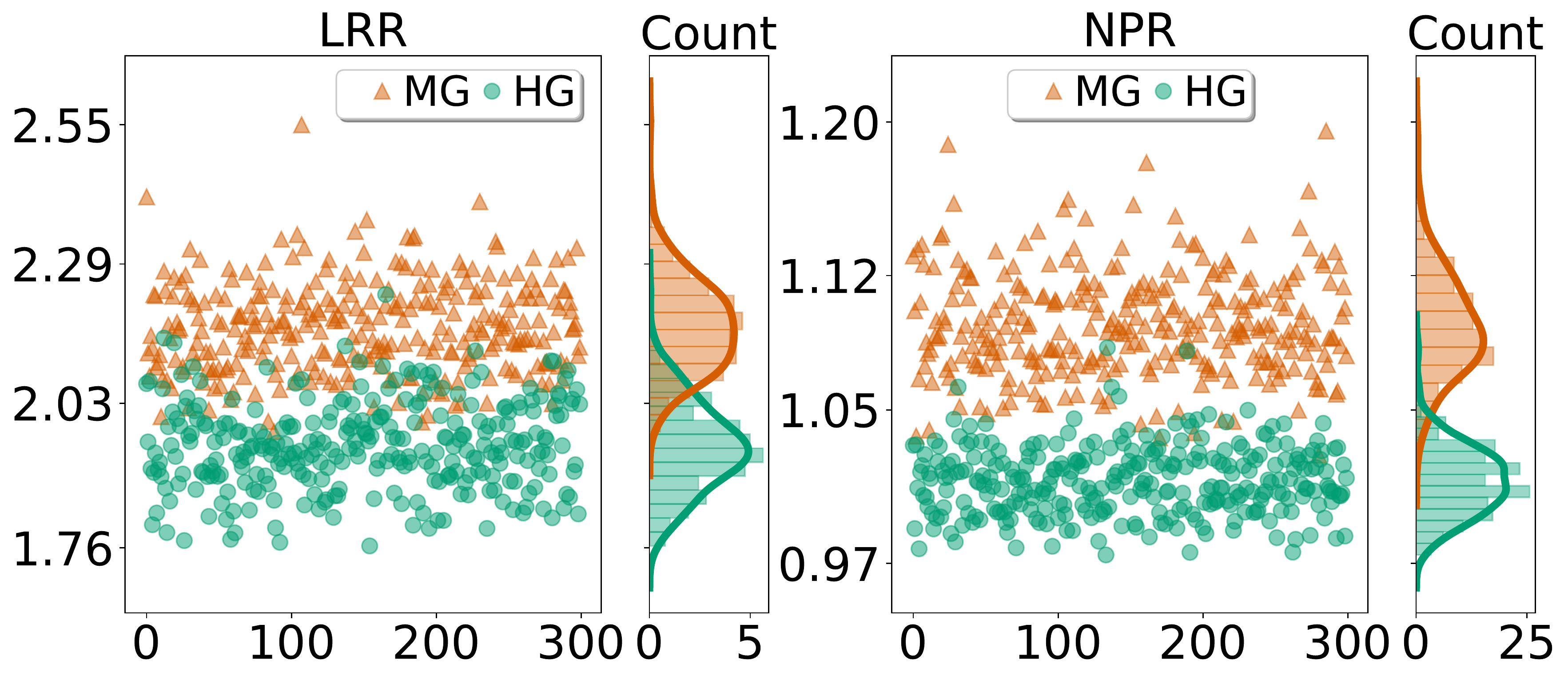}
    \caption{Distribution of \methodone and \methodtwo visualized on 300 human-written texts (HG) from the WritingPrompts dataset \cite{fan2018hierarchical} as well as 300 texts generated with GPT2-xl (MG) by prompting it with the first 30 tokens from human-written texts.}
    \label{fig: Intuition}
\end{figure}

  
\section{Related Work}
The detection of machine-generated text is commonly formulated as a classification task \cite{jawahar2020automatic,fagni2021tweepfake,bakhtin2019real,sadasivan2023can}. One way of solving it is to use supervised learning, where a classification model is trained on a dataset containing both machine-generated and human-written texts. For example, GPT2 Detector \cite{solaiman2019release} fine-tunes RoBERTa \cite{liu2019roberta} on the output of GPT2, while the ChatGPT Detector \cite{guo2023close} fine-tunes RoBERTa on the HC3 \cite{guo2023close} dataset.
However, models trained explicitly to detect machine-generated texts may overfit to their training distribution of the domains \cite{bakhtin2019real, uchendu2020authorship}.

Another stream of work attempts to distinguish machine-generated from human-written texts based on statistical irregularities in the entropy \cite{lavergne2008detecting}, perplexity \cite{beresneva2016computer} or in the $n$-gram frequencies \cite{badaskar2008identifying}.
\citet{gehrmann2019gltr} introduced hand-crafted statistical features to assist humans in detecting-machine generated texts. Moreover, \cite{solaiman2019release} noted the efficacy of simple zero-shot methods for detecting machine-generated text by evaluating the per-token log probability of texts and using thresholding. \citet{mitchell2023detectgpt} observed that machine-generated texts tend to lie in the local curvature of the log probability and proposed DetectGPT, whose prominent performance can only be guaranteed by the large size of the perturbation function and by a large number of perturbations, and thus costs more computational resources.

Other work explored watermarking, which imprints specific patterns of the LLM output text that can be detected by an algorithm while being imperceptible to humans. \citet{grinbaum2022ethical} and \citet{abdelnabi2021adversarial} watermarked machine-generated text using syntax tree manipulation, while \citet{kirchenbauer2023watermark} required access to the LLM's logits at each time step to add the watermark. 

\section{Improved Zero-Shot Approaches by Leveraging Log Rank Information}

\begin{figure}[t!]
    \centering
    \includegraphics[width=0.5\textwidth]{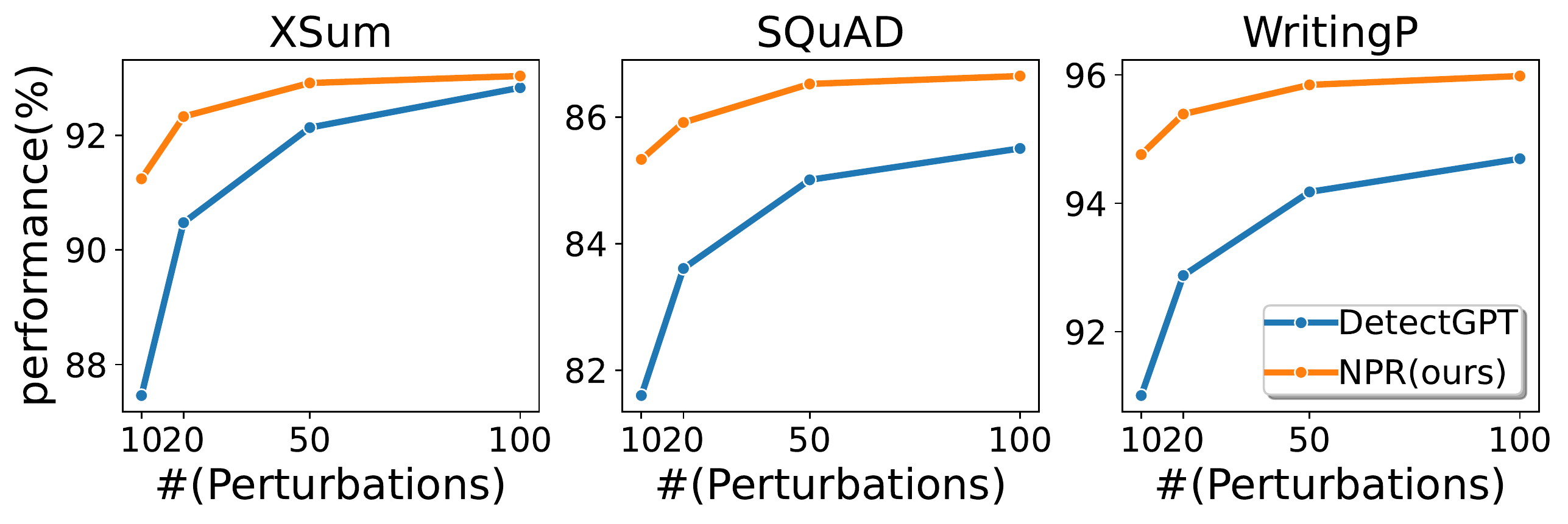}
    \caption{Comparison of DetectGPT to \methodtwo averaged across six models (in terms of AUROC). (The full results are given in Figure~\ref{fig: rank vs prob- 6 models} in the Appendix).}
    \label{fig: rank vs prob-average}
\end{figure}

In this section, we introduce the Log-\textbf{L}ikelihood Log-\textbf{R}ank \textbf{R}atio (\methodone) and the \textbf{N}ormalized \textbf{P}erturbed log-\textbf{R}ank (\methodtwo). \methodone combines log-rank and log-likelihood as they provide complimentary information about the evaluated text. \methodtwo is based on the idea that the log rank of machine-generated texts should be more sensitive to smaller perturbations.

\subsection{Log-Likelihood Log-Rank Ratio (\methodone)}

We define Log-Likelihood Log-Rank Ratio as
\begin{align*}
\text{\methodone} = &\left| \frac{\frac{1}{t}\sum_{i=1}^t\log p_{\theta}(x_i|x_{<i})}{\frac{1}{t}\sum_{i=1}^t\log r_{\theta}(x_i|x_{<i})}\right|\\ =& - \frac{\sum_{i=1}^t\log p_{\theta}(x_i|x_{<i})}{\sum_{i=1}^t\log r_{\theta}(x_i|x_{<i})},
\end{align*}
where $r_{\theta}(x_i|x_{<i})\geq 1$ is the rank of token $x_i$ conditioned on the previous tokens.

The Log-Likelihood in the numerator represents the absolute confidence for the correct token, while the Log-Rank in the denominator accounts for the relative confidence, which reveals complimentary information about the texts.  As illustrated in Figure~\ref{fig: Intuition}, \methodone is generally larger for machine-generated text, which can be used for distinguishing machine-generated from human-written text. One plausible reason might be that for machine-generated text, the log rank is more discernible than the log likelihood, so \methodone illustrates this pattern for machine-generated text. In Sections \ref{sec: exp} and \ref{sec: more exp}, we experimentally show that \methodone is a better discriminator than either the log likelihood or the log rank. We call the zero-shot method using \methodone as a detection feature as \methodonelong, and use abbreviation \methodone in the rest of the paper.

\subsection{Normalized Log-Rank Perturbation (\methodtwo)}

We define the normalized perturbed log rank as
\begin{equation*}
\text{\methodtwo} = \frac{\frac{1}{n}\sum_{p=1}^n\log r_{\theta}(\tilde{x}_p)}{\log r_{\theta}(x)},
\end{equation*}
where small perturbations are applied on the target text $x$ to produce the perturbed text $\tilde{x}_p$. Here, perturbation means minor rewrites of the texts, such as replacing some of the words. We call the zero-shot method using \methodtwo as a detection feature as \methodtwolong, and use abbreviation \methodtwo in the rest of the paper.

The motivation for \methodtwo is that machine-generated and human-written texts are both negatively affected by small perturbations, i.e., the log rank score will increase after perturbations, but machine-generated text is more susceptible to perturbations and thus increasing more on log rank score after perturbation, which suggests higher \methodtwo score for machine-generated texts. As shown in Figure~\ref{fig: Intuition}, \methodtwo can be a discernible signal for distinguishing machine-generated from human-written text. 
DetectGPT \cite{mitchell2023detectgpt} uses a similar idea, but experimentally, we find \methodtwo to be more efficient and to perform better. Details and comparisons are given in Section~\ref{sec: exp}.

\section{Experimental Setup}\label{sec: exp}

In this section, we conduct comprehensive experiments to evaluate the performance of \methodone and \methodtwo in comparison to several methods previosuly proposed in the literatu. 
We experiment with LLMs sizes varying from 1.5B to 20B parameters, probing the boundary of zero-shot methods when LLMs continue to grow in size. We further study the impact of the perturbation function used, the number of perturbations (specially for \methodtwo and DetectGPT), the decoding strategy, and the temperature used.

\subsection{Data}

Following \cite{mitchell2023detectgpt}, we use three datasets: XSum \cite{narayan2018don}, SQuAD \cite{rajpurkar2016squad}, WritingPrompts \cite{fan2018hierarchical}, containing news articles, Wikipedia paragraphs and prompted stories, respectively as human-written texts while attain machine-generated texts using LLMs. These datasets are chosen to represent the areas where LLMs could have a negative impact. For each experiment, we evaluate 300 machine-generated and human-written texts pairs by prompting the LLMs with the first 30 tokens of the human-written text. The whole data-generation process is released with our codes. 

\subsection{Evaluation Measure}

Following previous work \cite{mitchell2023detectgpt,he2023mgtbench,krishna2023paraphrasing}, we measure the performance using the area under the receiver operating characteristic curve (AUROC), which is the probability that a classifier correctly ranks the machine-generated example higher than human-written example. Since for zero-shot methods, detection rates are heavily dependent on the threshold when using discriminative statistics, the AUROC metric is commonly used to measure zero-shot detector performance, which considers the range of all possible thresholds \cite{krishna2023paraphrasing}.

\begin{table*}[t!]
\centering
\setlength{\tabcolsep}{1.5pt} 
\begin{tabular}{p{1.3cm}p{1.8cm}p{1.8cm}p{1.35cm}p{1.35cm}p{1.35cm}p{1.15cm}p{1.35cm}p{1.5cm}p{1.1cm}p{1.1cm}}
\toprule
Dataset&Perturbation& Method& GPT2-xl& Neo-2.7& OPT-2.7&GPT-j&OPT-13&Llama-13& NeoX&Avg.\\
\hline
\multirow{7}{*}{XSum}&\multirow{5}{*}{w/o}&$\log p$ &
89.16&
87.69&
86.98&
83.10&
83.90&
56.89&
78.16&
80.84 \\

&&Rank&
79.79&
77.87&
76.07&
76.28&
74.10&
48.81&
72.44&
72.19 \\

&&Log Rank&
91.75&
90.79&
\textbf{89.18}&
86.42&
\textbf{85.88}&
61.33&
81.44&
83.83 \\

&&Entropy&
56.78&
55.14&
50.34&
55.51&
50.98&
69.43&
60.84&
57.00 \\

&&\methodone(ours)&
\textbf{93.47}&
\textbf{92.24}&
88.70&
\textbf{88.68}&
83.79&
\textbf{71.07}&
\textbf{83.89}&
\textbf{85.98} \\

\cline{2-11}&\multirow{2}{*}{w/}&DetectGPT&
98.80&
99.11&
96.02&
\textbf{95.88}&
92.65&
73.55&
93.58&
92.80 \\

&&\methodtwo(ours)&
\textbf{99.40}&
\textbf{99.46}&
\textbf{97.09}&
95.76&
\textbf{94.63}&
\textbf{75.51}&
\textbf{94.08}&
\textbf{93.70} \\ 
\hline\hline
\multirow{7}{*}{SQuAD}&\multirow{5}{*}{w/o}&$\log p$ &
90.72&
84.18&
87.84&
78.20&
80.65&
42.91&
68.78&
76.18 \\

&&Rank&
83.46&
79.77&
81.85&
79.46&
77.47&
54.44&
73.10&
75.65 \\

&&Log Rank&
94.33&
89.52&
91.76&
83.37&
85.05&
48.28&
73.88&
80.88 \\

&&Entropy&
57.97&
58.48&
53.29&
58.26&
57.14&
\textbf{69.71}&
59.97&
59.26 \\

&&\methodone(ours)&
\textbf{97.42}&
\textbf{95.74}&
\textbf{95.89}&
\textbf{91.59}&
\textbf{91.36}&
68.78&
\textbf{83.31}&
\textbf{89.15} \\

\cline{2-11}&\multirow{2}{*}{w/}&DetectGPT&
98.52&
95.86&
96.91&
88.66&
90.60&
47.03&
76.84&
84.92 \\

&&\methodtwo(ours)&
\textbf{99.40}&
\textbf{97.56}&
\textbf{98.39}&
\textbf{91.88}&
\textbf{93.04}&
\textbf{48.67}&
\textbf{79.73}&
\textbf{86.95} \\ 

\hline\hline

\multirow{7}{*}{WritingP}&\multirow{5}{*}{w/o}&$\log p$ &
96.71&
95.63&
95.05&
94.43&
92.53&
83.54&
93.27&
93.02 \\

&&Rank&
87.62&
82.79&
83.89&
83.21&
83.52&
77.64&
81.64&
82.90 \\

&&Log Rank&
98.02&
97.15&
96.32&
96.06&
94.34&
88.11&
95.14&
95.02 \\

&&Entropy&
36.45&
34.07&
39.75&
36.93&
42.49&
47.64&
37.89&
39.32 \\

&&\methodone(ours)&
\textbf{98.34}&
\textbf{98.02}&
\textbf{96.45}&
\textbf{96.97}&
\textbf{95.09}&
\textbf{92.66}&
\textbf{96.56}&
\textbf{96.30} \\

\cline{2-11}&\multirow{2}{*}{w/}&DetectGPT&
99.30&
98.71&
98.33&
95.52&
96.46&
83.01&
92.94&
94.90 \\

&&\methodtwo(ours)&
\textbf{99.78}&
\textbf{99.59}&
\textbf{98.87}&
\textbf{98.07}&
\textbf{98.14}&
\textbf{89.39}&
\textbf{96.72}&
\textbf{97.22} \\ 
\bottomrule
\end{tabular}
\caption{\textbf{Zero-shot experiments.} Comparison of the proposed \methodone and \methodtwo to other zero-shot methods in terms of AUROC. For fair comparison, we show in bold the best results, both with and without perturbations.}
\label{tab:main}
\end{table*}

\subsection{Methods}

\subsubsection{Zero-Shot Methods}

We compare the following zero-shot methods:

\begin{itemize}
    \item \textbf{$\log p(x)$:} the idea is that a passage with a high average log probability is more likely to have been generated by the target LLM;
    \item \textbf{Rank:} the idea is that a passage with a higher average rank is more likely to have been generated by the target LLM;
    \item \textbf{Log-Rank:} passage with higher average observed log rank is more likely to have been generated by the target LLM;
    \item \textbf{Entropy:} machine-generated text has higher entropy;
    \item \textbf{DetectGPT:} machine-generated text has more negative log probability curvature.
\end{itemize}

More detail and exact definition of these methods can be found in Appendix~\ref{Appe: baselines}.

These zero-shot baselines, along with our newly proposed \methodone and \methodtwo, can be categorized into two groups:
\begin{itemize}
\item \textbf{Perturbation-free}: $\log p(x)$, Rank, Log-Rank, Entropy, \methodone.
These methods only query the LLM for statistics about the target text $x$.
\item \textbf{Perturbation Based}: DetectGPT and \methodtwo. 
These methods query the LLM not only for the target text $x$, but also for perturbed versions thereof $\tilde{x}_1,\cdots, \tilde{x}_p$. 
\end{itemize}

Since perturbation-based methods generally perform better (but are also more time-consuming), for fair comparison, we compare them within their own group.

\subsubsection{Supervised Methods}

We also experiment with two supervised detectors: RoBERTa-base and RoBERTa-Large. However, as these are not central for our narrative, we put the results and the analysis in Appendix~\ref{Appe: supervised}.

\subsection{Experimental Details} 

For the perturbation-based methods (DetectGPT and \methodtwo), we use T5-3B as the perturbation model and we perturb the input text 50 times for all the experiments, unless specified otherwise. For all zero-shot methods, we use sampling with a temperature of 1, unless specified otherwise. More details about the experiments are given in Appendix~\ref{Appe: baselines}.

\begin{table*}[t!]
\centering
\begin{tabular}{lllllll|ll}
\toprule
&&\multicolumn{5}{c|}{w/o Perturbation}&\multicolumn{2}{c}{w/ Perturbation}\\
\hline
Decoding&Dataset&$\log p$ &
                      Rank&
                      Log Rank&
                    Entropy&
                    \methodone(ours)
                  &DetectGPT&
                   \methodtwo(ours)\\
\hline
\multirow{3}{*}{top-$k$}
&XSum&
81.64&
70.68&
85.19&
55.47&
\textbf{89.25}&
91.34&
\textbf{92.93} \\

&SQuAD&
76.31&
74.31&
81.28&
57.96&
\textbf{90.61}&
82.42&
\textbf{84.99} \\

&WritingP&
93.80&
82.15&
95.72&
37.26&
\textbf{97.10}&
93.89&
\textbf{96.33} \\ 
\midrule

\multirow{3}{*}{top-$p$}
&XSum&
86.94&
70.86&
\textbf{88.65}&
53.89&
88.29&
92.74&
\textbf{93.42} \\

&SQuAD&
82.07&
75.03&
85.49&
55.86&
\textbf{91.09}&
83.98&
\textbf{86.19} \\

&WritingP&
96.51&
82.48&
\textbf{97.44}&
33.92&
97.25&
94.20&
\textbf{96.55} \\ 
\bottomrule
\end{tabular}
\caption{\textbf{Decoding strategy analysis.} Shown are the AUROC scores for methods with top-$k$ ($k=40$) and top-$p$ ($p=0.96$) sampling averaged across four LLMs: Neo-2.7, OPT-2.7, GPT-j, Llama-13.}
\label{tab:Average-Diff-decoding}
\end{table*}

\section{Evaluation Results}
\label{sec:results}

\subsection{Zero-Shot Results}

Table \ref{tab:main} shows a comparison of the five baseline zero-shot approaches to our proposed \methodone and \methodtwo, grouped as perturbation-based and perturbation-free. We can see that for the perturbation based methods, \methodtwo consistently outperforms DetectGPT on all datasets and LLMs, except for one case, with an average improvement of 0.90, 2.03, 2.32 AUROC points absolute on XSum, SQuAD, and WritingPrompts, respectively, (using the same perturbation function and the same number of perturbations). For the experiments among perturbation-free methods, on average, our method achieves the best performance and improves by 2.15, 8.27, 1.28 AUROC points absolute over the second-best perturbation-free method (i.e., log rank) on XSum, SQuAD, and WritingPrompts, respectively. 
Moreover, we find that in some cases, \methodone can even perform  better than perturbation based methods, e.g.,~on SQuAD, \methodone outperforms DetectGPT by 4.23 AUROC point absolute and outperforms \methodtwo by 2.20 AUROC points.

\subsection{Comparing DetectGPT to \methodtwo}\label{sec: compare DetectGPT}
Equipped with large perturbation functions and adequate amount of perturbations, perturbation-based methods generally outperform perturbation-free ones, e.g.,~using T5-3b as the perturbation function and perturb 50 times as in Table \ref{tab:main}. However, in practice, due to time and resource constraint, not all users can afford these models and large amount of perturbations. Thus, it is important to investigate how \methodtwo and DetectGPT behave with smaller perturbation function size and fewer perturbations.

\paragraph{Different Number of Perturbations.} Figure \ref{fig: rank vs prob-average} shows the averaged performance of DetectGPT and \methodtwo with varying number of perturbations. We can see that \methodtwo consistently performs better than DetectGPT when using the same number of perturbations. In other words, \methodtwo can achieve a comparable or better performance but with significantly fewer perturbations. For example, in SQuAD and WritingPrompts dataset, \methodtwo achieves 85 points and 95 points using approximately 10 perturbations while DetectGPT requires around 100 perturbations, which  highlights the effectiveness and efficiency of \methodtwo. More complete results for each dataset and models can be found in Figure \ref{fig: rank vs prob- 6 models} and Figure \ref{fig: rank vs prob- neoX-20b} of Appendix \ref{Appe: DetectGPT vs methodtwo}.

\paragraph{Different Perturbation Functions.} 
In Table \ref{tab:Average-different perturbation function}, we compare \methodtwo with DetectGPT using smaller perturbation model T5-large, and the result is averaged over 6 LLMs and 3 datasets. We found that, not surprisingly, replacing T5-3b to smaller models harms the performance of both \methodtwo and DetectGPT,  and the performance degradation can't be mitigated by increasing the number of perturbations.  For both \methodtwo and DetectGPT, the average performance of 100 perturbations with T5-large is still worse than 10 perturbations with T5-3b (emphasized with gray box in Table \ref{tab:Average-different perturbation function}). Moreover, one can observe that, \methodtwo is less affected by the reduced perturbation function size: when replacing T5-3b to T5-large, the performance degradation  averaged over 10, 20, 50, 100 perturbations for \methodtwo is 4.40 point, much smaller compared to that of 8.06 point  for DetectGPT. The complete results on 6 LLMs and 3 datasets can be found in Figure \ref{fig: rank vs prob- t5-large} of Appendix  \ref{Appe: DetectGPT vs methodtwo}.





\begin{table}[t!]
\centering
\begin{tabular}{p{1.4cm}P{1.7cm}P{0.6cm}P{0.6cm}P{0.6cm}P{0.6cm}}

\toprule
\multirow{2}{*}{
{\splitcell{Perturbation\\Function}} }
&\multirow{2}{*}{
Dataset
}&\multicolumn{4}{c}{\# (Perturbations)}\\\cline{3-6}
& &10 &20&50&100\\\hline

\multirow{3}{*}{
\methodtwo (ours)
}
&T5-large
& 86.69
& 88.00
& 88.74
&  \cellcolor[gray]{0.8}{88.94}
\\\cline{3-6}
&T5-3b
& \cellcolor[gray]{0.8}{91.39}
& 92.35
& 93.04
& 93.20
\\\cline{3-6}
&Diff
& \textbf{4.70}
& \textbf{4.35}
& \textbf{4.30}
& \textbf{4.26}
\\\hline\hline

\multirow{3}{*}{
DetectGPT
}
&T5-large
& 77.94
& 81.12
& 83.90
& \cellcolor[gray]{0.8}{84.54}
\\\cline{3-6}
&T5-3b
& \cellcolor[gray]{0.8}{86.70}
& 89.57
& 91.38
& 92.10
\\\cline{3-6}
&Diff
& 8.76
& 8.45
& 7.48
& 7.56
\\\bottomrule

\end{tabular}
\caption{\textbf{Perturbation analysis.} Comparing DetectGPT to \methodtwo using different perturbations (AUROC scores).}
\label{tab:Average-different perturbation function}
\end{table}

\subsection{Different Decoding Strategy and Temperature}

In this subsection, we study different decoding strategy and temperature to see how these factors impact different zero-shot detectors.
\begin{table*}[t!]
\centering
\begin{tabular}{p{1.8cm}|p{1.2cm}p{1.2cm}p{1.6cm}p{1.2cm}p{1.8cm}|p{1.8cm}p{1.8cm}}
\toprule
&\multicolumn{5}{c}{w/o Perturbation}&\multicolumn{2}{c}{w/ Perturbation}\\
\hline
Temperature&$\log p$ &
                      Rank&
                      Log Rank&
                    Entropy&
                    \methodone(ours)
                  &DetectGPT&
                   \methodtwo(ours)\\
                   \hline
0.5&
98.72&
77.87&
\textbf{
99.29
}
&
25.90&
99.23&
86.14&
\textbf{
95.76
}
\\\hline
0.7&
97.01&
76.98&
98.05&
38.28&
\textbf{
98.84
}
&
90.28&
\textbf{
95.61
}
\\\hline
0.9&
90.04&
75.82&
92.28&
47.14&
\textbf{
94.50
}
&
90.33&
\textbf{
92.89
}
\\\hline
0.95&
86.15&
75.43&
88.88&
50.42&
\textbf{
92.04
}
&
89.97&
\textbf{
91.98
}
\\\hline
1&
81.48&
74.85&
84.81&
52.37&
\textbf{
89.15
}
&
89.02&
\textbf{
90.86
}
\\\bottomrule
\end{tabular}
\caption{\textbf{Temperature experiments.} Results of using different temperatures (AUROC scores).}
\label{tab: different temperature}
\end{table*}

\paragraph{Alternative Decoding Strategies.}
In line with prior work \cite{pagnoni2022threat}, we experimented with top-$k$ sampling \cite{fan2018hierarchical} and top-$p$ sampling \cite{holtzman2019curious}. Top-$k$ sampling generates from top-$k$ most likely words according to the LLM. Top-$p$ sampling (nucleus sampling) samples from the set of words that collectively accounts for a total mass probability $p$. The result (averaged across 4 LLMs) are shown in Table \ref{tab:Average-Diff-decoding}, and complete results can be found in Table \ref{tab: top-k, top-p} of Appendix \ref{Appe: sampling strategy and temperature}. We find that, although almost all the zero-shot methods performs better when using top-$k$ and top-$p$ sampling than temperature sampling, Log Rank and Log Likelihood method are more in favor of top-$p$ sampling, while \methodone is stable in both top-$p$ and top-$k$ sampling. 
For top-$k$ decoding, \methodone improves 4.06, 9.33, 1.38 points over the second best zero-shot method baseline on three datasets, respectively. \methodone performance also improves when using top-$p$ decoding strategy, but due to the unstable performance surge of Log Rank method, \methodone
become slightly behind Log Rank method, with a minor difference of 0.36 and 0.19 points on XSum and WritingPrompts dataset, respectively. 
For perturbation based methods, their behavior is consistent with previous results, where \methodtwo outperforms DetectGPT for both top-$p$ and top-$k$ sampling strategies.

\paragraph{Different Temperature.} Temperature controls the degree of randomness of the generation process. Increasing the temperature leads to more randomness and creativity, while reducing it leads to more conservation and less novelty. In practice, people adjust temperature for their specific purposes. For example, students might set a high temperature to encourage more original and diverse output when writing a creative essay, whereas fake news producers might set lower temperatures to generate seemingly convincing news articles for their deceptive purposes. Based on our experiments in Table \ref{tab: different temperature}, we found that Log Likelihood ($\log p$), Log Rank and \methodone is highly sensitive to the temperature and can get even better results than perturbation based methods when the temperature is relatively low. In addition, the performance improvement of Rank method with the increased temperature is negligible compared to Log Likelihood, Log Rank and \methodone, while the performance of entropy method seems to be positively correlated to the temperature. We conjure that the abnormal behavior of Entropy method might be because of the assumption ``machine generated text has higher entropy" \cite{mitchell2023detectgpt}, which, from our experiments, doesn't stand for high temperature. As for perturbation based method, the impact of temperature is not so clear as perturbation-free method. But in general, the results suggest
the temperature has only minor effects on DetectGPT while it improves the performance of \methodtwo. Another observation is that, perturbation-free method performs better than perturbation based method in low temperature, for example, for temperature is smaller than 0.95, perturbation based methods get better detection accuracy while being efficient.

\section{Analysis of the Efficiency}\label{sec: more exp}
\begingroup
\begin{table*}[t!]
\centering
\setlength{\tabcolsep}{3.5pt} 
\renewcommand{\arraystretch}{1} 
\begin{tabular}
{cccc|ccccccc}
\toprule
\multicolumn{4}{c}{$t_p(s)$}&\multicolumn{7}{c}{$t_m(s)$}\\
\hline
T5-3b&T5-large& T5-base&  T5-small& GPT2-xl& Neo-2.7& OPT-2.7&GPT-j&OPT-13&Llama-13& NeoX\\
\hline
0.10
&
0.08
&
0.04
&
0.03
&
0.06
&
0.09
&
0.10
&
0.04
&
0.07
&
0.07
&
0.60
\\\bottomrule

\end{tabular}
\caption{\textbf{Computation time.} Estimated computation time for one perturbation ($t_p$) and for calculating the target statistics on the text ($t_m$): shown in seconds.}
\label{tab:computational time-disassemble}
\end{table*}
\endgroup

Though in Table \ref{tab:main}, perturbation based methods appear to be significantly better than perturbation-free methods, it is important to note that their superior performances can only be achieved with large perturbation functions and multiple number of perturbations, which leads to intensive demand for computational resources and longer computational time. Thus, while performance is an important factor, it is crucial to consider the efficiency of these zero-shot methods as well.

\subsection{Computational Cost Analysis}
To get an idea of how costly different zero-shot methods are to achieve their performance in Table \ref{tab:main}, we estimated the computational time (per sample) for each zero-shot method (Detailed results can be found in Table \ref{tab:computational time} of Appendix \ref{appe: computational time}).
For perturbation based methods, we used 50 perturbations with T5-3b as in the main experiment for the estimation. We observed that the computational time of Log Likelihood, Rank, Log Rank and Entropy are almost the same, while \methodone runs approximately 2 times longer than these methods, since it requests both the Log Rank and Log Likelihood statistics. For perturbation based methods, the running time is at least 50 times longer compared to Log Likelihood, Rank, Log Rank, Entropy method, since they calculate the Log Likelihood or Log Rank for not only the target text, but also perturbed samples.

\paragraph{Composition of the Computational Time.} In general, for perturbation-free zero-shot methods, the computational time only depends on the size of LLM and the complexity of statistics. \methodone is twice as complex as simple statistics such as Log Rank and Log Likelihood, so it takes approximately twice as long to compute. As for LLM size, intuitively, larger models usually takes more time to compute, which can also be observed in Table \ref{tab:computational time}.
The additional computational time of perturbation based methods comes from two folds: (1) The total time for perturbation, which depends on the perturbation function we use and the number of perturbations. (2) The total time for calculating statistics of the perturbed texts, which depends on the number of perturbations, the size of LLM and the complexity of statistics. To reduce the computational time of perturbation based method, we could either choose smaller size of perturbation function or reduce the number of perturbations. 

\paragraph{Formula for Estimating the Computational Time.}
Let $t_p$ be the time of perturbing 1 sample, $t_m$ be the time of calculating a simple statistics (such as log likelihood) of one sample for a particular LLM and $n$ be the number of perturbations. The computational time for log likelihood, rank, log rank, entropy is approximately $t_m$, the estimated time for \methodone is $2\cdot t_m$, while the estimated computational time for perturbation based method is $n\cdot  t_p + (n+1)\cdot t_m$. The estimated value of $t_p$ and  $t_m$ are illustrated in Table \ref{tab:computational time-disassemble}, which can help us estimate the total running time (in seconds) of different zero-shot methods.

\subsection{Balancing Efficiency and Performance}
In this subsection, we provide additional experiments on \methodone (the best perturbation-free method) and \methodtwo (the best perturbation based method, more time consuming than \methodone but also rather satisfactory performance) to provide users some intuition on setting parameters of \methodtwo and choosing among between these two methods according to user's preference of efficiency and performance.

First, we study the perturbation function used for \methodtwo. Different from Section \ref{sec: compare DetectGPT}, where the focus is to illustrate the advanced performance of \methodtwo compared with DetectGPT, here, we mainly focus on the efficiency performance trade-off perspective and provide some intuition on choosing perturbation functions.

\begin{figure*}[t!]
    \centering
    \includegraphics[width=1\textwidth]{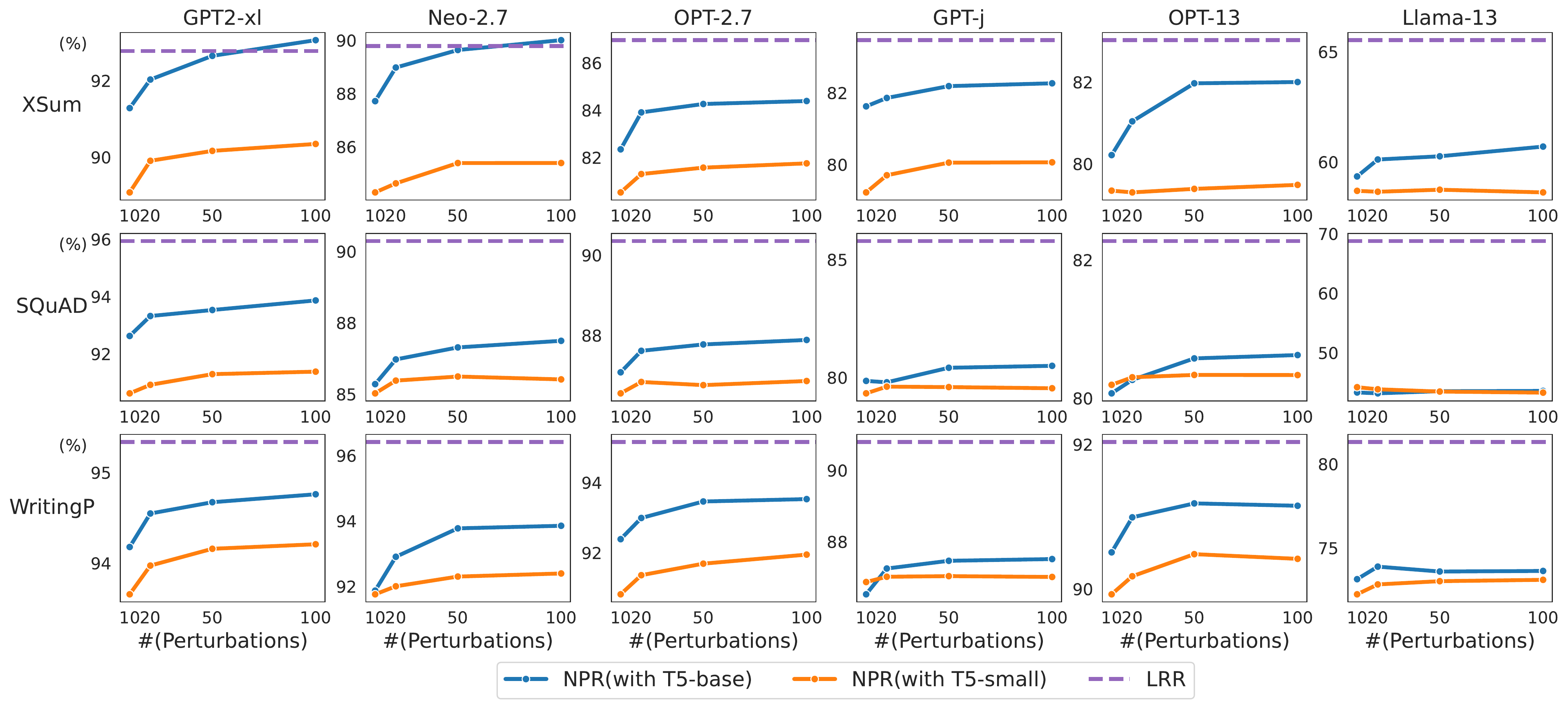}
    \caption{Comparing \methodone and \methodtwo when T5-small and T5-base are used for perturbation in \methodtwo  (AUROC scores).}
    \label{fig: t5-small and t5-base}
\end{figure*}

\paragraph{T5-small and T5-base are not good candidates for perturbation functions.} T5-small and T5-base are 2 or 3 times faster than larger models such as T5-large (as shown in Table \ref{tab:computational time-disassemble}), one might wonder if it is possible to trade the saved time with more perturbations for a better performance? We give a negative answer to this. We observe in Figure \ref{fig: t5-small and t5-base} that using T5-base and T5-small performs worse than \methodone even with 50 to 100 perturbations, which suggests that \methodone can be at least 50 to 100 times faster while outperform perturbation based methods. So, if the user can only afford T5-small or T5-base as perturbation function, they should choose \methodone with no hesitation since it achieves both better efficiency and better performance.
\begin{figure*}[t!]
    \centering
    \includegraphics[width=1\textwidth]{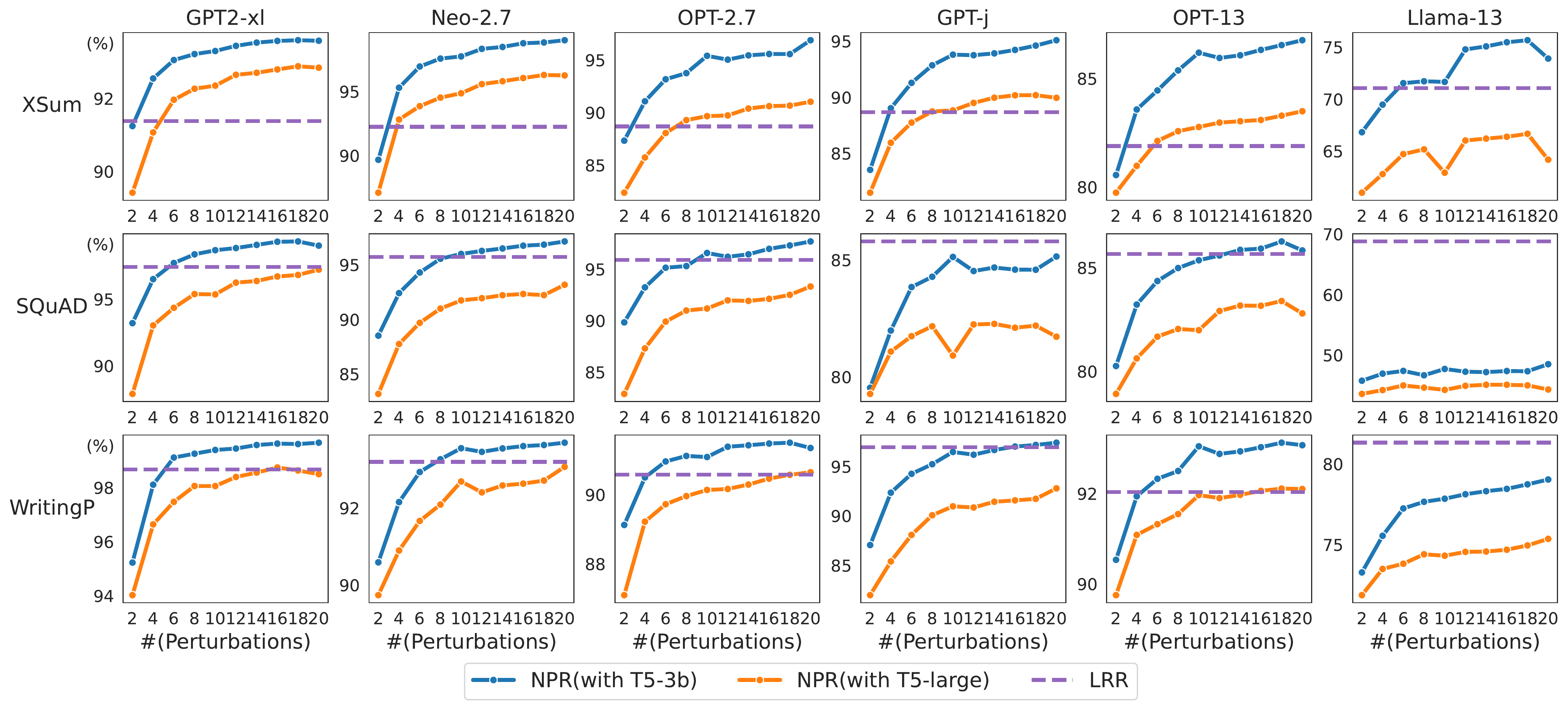}
    \caption{Comparing \methodone and \methodtwo when T5-3b and T5-large are used for the perturbation in \methodtwo (AUROC scores).}
    \label{fig: t5-large and t5-3b}
\end{figure*}

\paragraph{Cost-Effectiveness on More Perturbations and Larger Perturbation Function.} In Figure \ref{fig: t5-large and t5-3b}, we illustrate the effectiveness of \methodone compared to \methodtwo with T5-large and T5-3b as perturbation function respectively, from which, we find that (1) T5-3b has a higher performance upper limits compared with T5-large. So, if resources are allowed (enough memory and adequate perturbation time), t5-3b would be a better choice, especially for users that prioritize performance.
(2) To achieve the same performance as \methodone, generally we only need less than 10 perturbations using T5-3b as perturbation function. This estimate could help us choose whether to use \methodtwo or \methodone on validation set: setting the number of perturbation to be 10, if \methodone outperforms \methodtwo, we would suggest use \methodone, otherwise, \methodtwo would be a better option.
(3) To achieve the same performance, using T5-large takes more than 2 times perturbations than using T5-3b, while the perturbation time using T5-3b is less than twice of the time using T5-large, so using large perturbation functions such as T5-3b is much more efficient than using smaller ones such as T5-large. The only concern is the memory. 

In summary, when parameterizing perturbation function and the number of perturbations, we suggest using the lager perturbation functions if memory permits, which is more cost-effective: less time-consuming for achieving the same performance and has a high performance upper limit. In addition, setting the number of perturbation to be 10 would be a good threshold on the validation set to decide whether to use \methodtwo or \methodone.

\section{Conclusion}
In this paper, we proposed two simple but effective zero-shot machine generated text detection methods by leveraging the log rank information. The methods we proposed ---\methodone and \methodtwo---, achieve state-of-the-art performance within their respective category. In addition, we explored different settings such as decoding strategy and temperatures, as well as different perturbation functions and number of perturbations to better understand the advantages and the disadvantages of different zero-shot methods. Then, we analyzed the computational costs of these methods, and we provided guidance on balancing  efficiency and performance.

\section{Limitations and Future Directions}
One of the limitation of zero-shot methods is the white box assumption that we can have some statistics about the source model. This induces two problems: for close-source models (such as GPT-3), these statistics might not have been provided. Moreover, in practice, the detector might have to run the model locally to get the statistics for the purpose of detection, which requires that the detector have enough resources to use the LLM for inference. 

Based on the limitations of zero-shot methods, we consider weak supervised learning \cite{ratner2017snorkel} as an important direction for future work. Though many papers in MGTD assume knowing the source LLM where the text is generated from, in realistic, the source LLM might be unknown, so it is worth combining weak supervised learning as well as weak supervision sources (other LLMs at hand that might not be the target LLM) to weakly train a classifier. With the flexibility of the weak supervision sources, the limitations of our work could possibly be addressed: (1) Since the weak supervision sources do not have to be from the same target model, there is no need to assume that the target LLM is  known.  (2) Since the weak supervision sources are classifiers, we could only use statistics that are within reach, or even statistics from other open-source LLMs. (3) The weak supervision sources can be from smaller LLMs, rather than the target LLM, this relaxes the requirement for running an extremely large LLM locally.

\bibliography{anthology,custom}
\bibliographystyle{acl_natbib}
\appendix
\onecolumn
\newpage

\section{Experimental Details and Baselines}\label{Appe: baselines}
\paragraph{Details on Baselines.}
We mainly compare the proposed methods with zero-shots methods, 
which utilize the source model itself to extract distinguishable statistic features, including:
\begin{itemize}
\item Log Likelihood ($\log p$) \cite{solaiman2019release}: This approach evaluates the average token-wise log probability of the text and classifies text with higher Log Likelihood to be machine generated.
\item Rank \cite{gehrmann2019gltr}: This approach evaluates the average rank of each token of the text and classifies text with smaller average rank to be machine generated.

\item Log Rank \cite{mitchell2023detectgpt}: Instead of using the Rank score directly, this approach evaluates the average Log Rank of each token of the text and classifies text with smaller average Log Rank to be machine generated.
\item Entropy \cite{gehrmann2019gltr}: This approach is inspired by the hypothesis that machine generated texts are more likely to have over-confident (thus low entropy) predictive distributions. In practice, \cite{mitchell2023detectgpt} discovered that entropy to be positively correlated with passage fakeness, therefore, following their convention, we use high average entropy as a signal of machine generated text.

\item DetectGPT \cite{mitchell2023detectgpt}: DetectGPT is based on the hypothesis that when applying small perturbations to a passage $x$ and produce the perturbed text $\tilde{x}$, the quantity $\log p_{\theta}(x)-\log p_{\theta}(\tilde{x})$ is relatively larger for machine generated samples than human written one. In practice, the performance of this approach depends heavily on the external perturbation function and the number of perturbations.
\end{itemize}

\paragraph{Details on LLMs used.} We used 7 LLMs ranging from 1.5B parameters to 20B parameters in our main experiments. 
\begin{itemize}
\item GPT2-xl \cite{radford2019language} is the 1.5B parameter version of GPT2 trained on a dataset of 8 million web pages called WebText \cite{radford2019language}, whose objective is to predict the next word given previous words within the text. GPT2-xl surpasses many other language models trained on specific domain (such as books, news, Wikipedia) without using domain-specific training dataset. 
\item GPT-Neo-2.7B \cite{gpt-neo} was trained as an autoregressive language model on Pile \cite{gao2020pile} dataset with EleutherAI's replication of the GPT-3 architecture. 
\item OPT-2.7B and OPT-13B are two models among a 
 collection of decoder-only pre-trained transformers introduced in \cite{zhang2022opt}, with the performance roughly match GPT-3 of the same size.
 \item GPT-j-6B \cite{gpt-j}, which was also trained on Pile \cite{gao2020pile}, exhibits zero-shot performance roughly comparable to GPT-3 of comparable size. In addition, the performance gap from GPT-3 of similar size is closer than the GPT-Neo models.
 \item Llama-13b is the 13B parameter model from Llama models \cite{touvron2023llama}: a collection of models ranging from 7B to 65B parameters trained with publicly available dataset. Llama-13B outperforms GPT-3
(175B) on most benchmarks, and all the models are released to the research community.
 \item NeoX-20B \cite{black2022gpt} is a 20B autoregressive model trained on Pile, whose weights have been released openly to the public.
\end{itemize}

\paragraph{Experimental Details.} For small models such as GPT2-xl, Neo-2.7, OPT-2.7, GPT-j, we use 1 NVIDIA A100 GPU (with total memory 40G) in our experiments; for larger models such as OPT-13b and Llama-13, we use 3 A100 GPUs (total memory 120 G) while using 4 A100 GPUs (total memory 160 G) for the largest model NeoX-20.

\section{Supervised Methods}\label{Appe: supervised}

\paragraph{Main results for supervised methods.}
Comparing Table \ref{tab:main} with Table \ref{tab:supervised}, we found that, on average, our best zero shot method (either \methodone on SQuAD dataset or \methodtwo on XSum and WritingPrompts dataset) can exceed supervised model fine-tuned on roberta-base. For larger model roberta-large, only on writing dataset, perturbation-based method DetectGPT and \methodtwo outperforms roberta-large model, by a margin of 0.55\% and 2.87\% respectively.

\paragraph{Supervised Method with Different Decoding Strategy.}
We experimented the 4 models used in zero-shot methods with top-$p$ and top-$k$ decoding strategy for supervised method and found that using top-$p$ decoding strategy performs better than using top-$k$. (See Table \ref{tab:supervised-topk-topp}). Compared to zero-shot methods, the best zero-shot method \methodtwo can outperform roberta-base model while being comparable to roberta-large model.
\begin{table*}[t!]
\centering
\setlength{\tabcolsep}{2pt} 
\begin{tabular}{p{1.4cm}p{2.2cm}p{1.35cm}p{1.35cm}p{1.35cm}p{1.15cm}p{1.35cm}p{1.5cm}p{1.1cm}p{1.1cm}}
\toprule
& &GPT2-xl& Neo-2.7& OPT-2.7&GPT-j&OPT-13&Llama-13& NeoX& Avg.\\
\hline

\multirow{2}{*}{XSum}&roberta-base&
97.57&
96.82&
94.86&
90.37&
88.62&
79.18&
88.96&
90.91 \\

&roberta-large&
\textbf{99.74}&
\textbf{99.73}&
\textbf{98.37}&
\textbf{97.58}&
\textbf{93.85}&
\textbf{85.93}&
\textbf{95.13}&
\textbf{95.76} \\ 

 \hline\hline
\multirow{2}{*}{SQuAD}&roberta-base&
97.65&
94.42&
92.56&
87.57&
88.96&
76.98&
84.37&
88.93 \\

&roberta-large&
\textbf{99.01}&
\textbf{98.30}&
\textbf{96.53}&
\textbf{93.31}&
\textbf{91.62}&
\textbf{82.59}&
\textbf{88.37}&
\textbf{92.82} \\ 

 \hline\hline

\multirow{2}{*}{WritingP}&roberta-base&
96.88&
95.23&
89.57&
93.26&
86.18&
83.49&
88.92&
90.50 \\

&roberta-large&
\textbf{98.75}&
\textbf{98.80}&
\textbf{94.98}&
\textbf{97.11}&
\textbf{88.75}&
\textbf{88.32}&
\textbf{93.72}&
\textbf{94.35} \\ 

\bottomrule

\end{tabular}
\caption{Complete results for the supervised methods (AUROC score).}
\label{tab:supervised}
\end{table*}
\begin{table*}[t!]
\centering
\setlength{\tabcolsep}{2.7pt} 
\begin{tabular}{p{1.4cm}p{2.0cm}p{1.35cm}p{1.35cm}p{1.2cm}p{1.5cm}|p{1.35cm}p{1.35cm}p{1.2cm}p{1.5cm}}
\toprule
&&\multicolumn{4}{c}{top-$k$}&\multicolumn{4}{c}{top-$p$}\\
\hline
& & Neo-2.7&OPT-2.7 & GPT-j&Llama-13&Neo-2.7&OPT-2.7 & GPT-j&Llama-13\\

\hline

\multirow{2}{*}{XSum}&roberta-base&
96.48&
94.15&
92.72&
82.47&
98.30&
97.30&
96.71&
85.84 \\

&roberta-large&
\textbf{99.74}&
\textbf{98.06}&
\textbf{98.29}&
\textbf{87.33}&
\textbf{99.84}&
\textbf{98.97}&
\textbf{98.82}&
\textbf{89.35} \\

 \hline\hline
\multirow{2}{*}{SQuAD}&roberta-base&
93.55&
93.27&
87.60&
76.79&
96.34&
97.65&
92.26&
84.05 \\

&roberta-large&
\textbf{98.35}&
\textbf{96.88}&
\textbf{93.97}&
\textbf{82.42}&
\textbf{98.21}&
\textbf{98.39}&
\textbf{95.09}&
\textbf{86.46} \\

 \hline\hline

\multirow{2}{*}{WritingP}&roberta-base&
97.27&
90.14&
93.86&
83.24&
98.33&
94.09&
96.55&
88.78 \\

&roberta-large&
\textbf{99.34}&
\textbf{96.34}&
\textbf{97.12}&
\textbf{87.05}&
\textbf{99.68}&
\textbf{96.06}&
\textbf{97.94}&
\textbf{89.59} \\ 

\bottomrule

\end{tabular}
\caption{Complete results for the supervised methods using top-$k$ ($k=40$) and top-$p$ ($p=0.96$) sampling across four models (AUROC scores).}
\label{tab:supervised-topk-topp}
\end{table*}

\begin{figure*}[t!]
    \centering
    \includegraphics[width=1\textwidth]{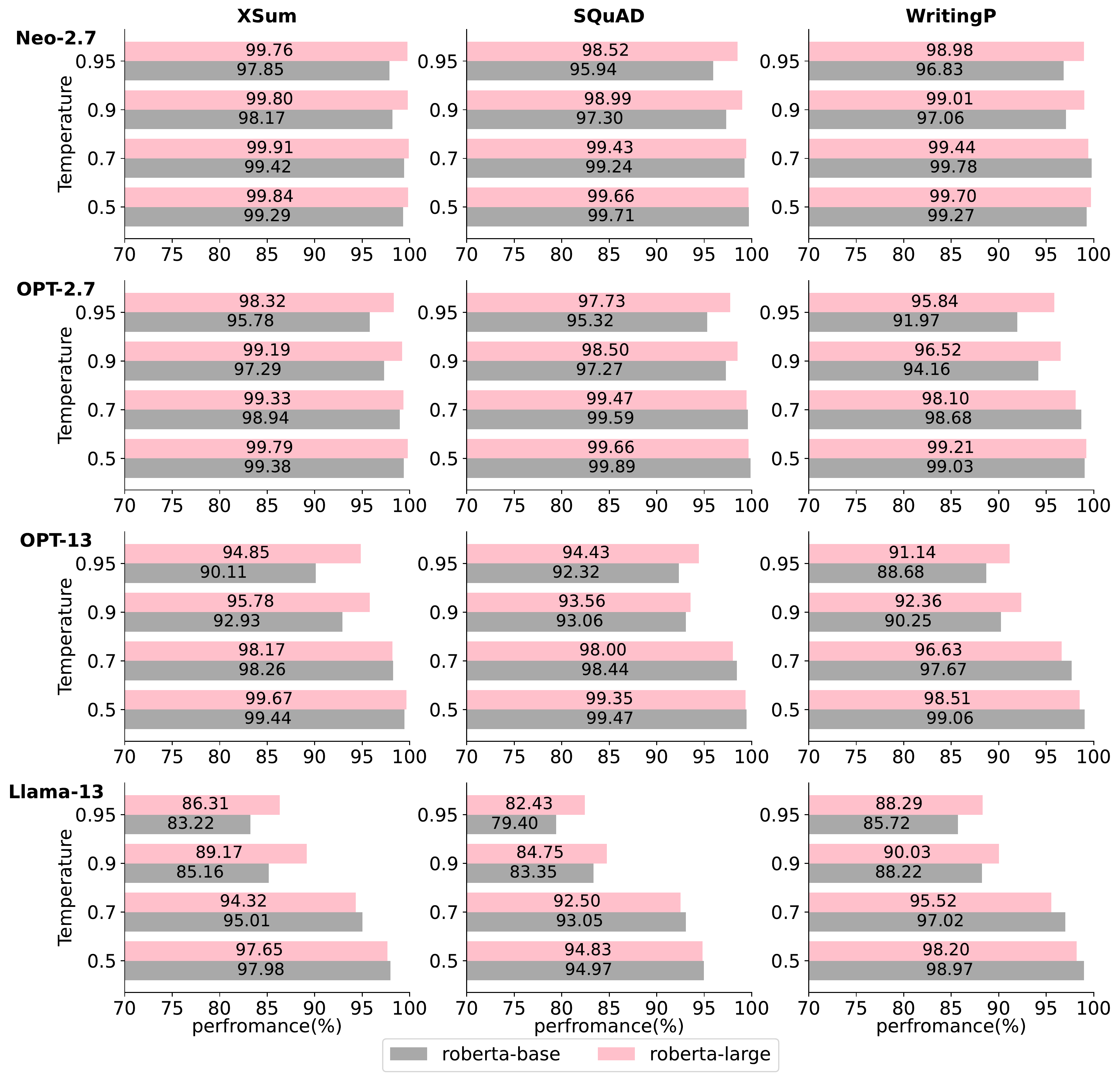}
    \caption{Comparing supervised methods with different temperature (AUROC score).}
    \label{fig: supervised-temperature}
\end{figure*}

\paragraph{Supervised Method with Different Temperature.} Supervised methods also perform better with lower temperature, but zero-shot methods such as Log Rank and Log Likelihood methods might exceed supervised methods in low temperature. Moreover, we found that the performance gap of roberta-base and roberta-large would be narrowed with lower temperature. The results are illustrated in Figure \ref{fig: supervised-temperature}.

\section{Comparing \methodtwo and DetectGPT}\label{Appe: DetectGPT vs methodtwo}
\begin{figure*}[t!]
    \centering
    \includegraphics[width=1\textwidth]{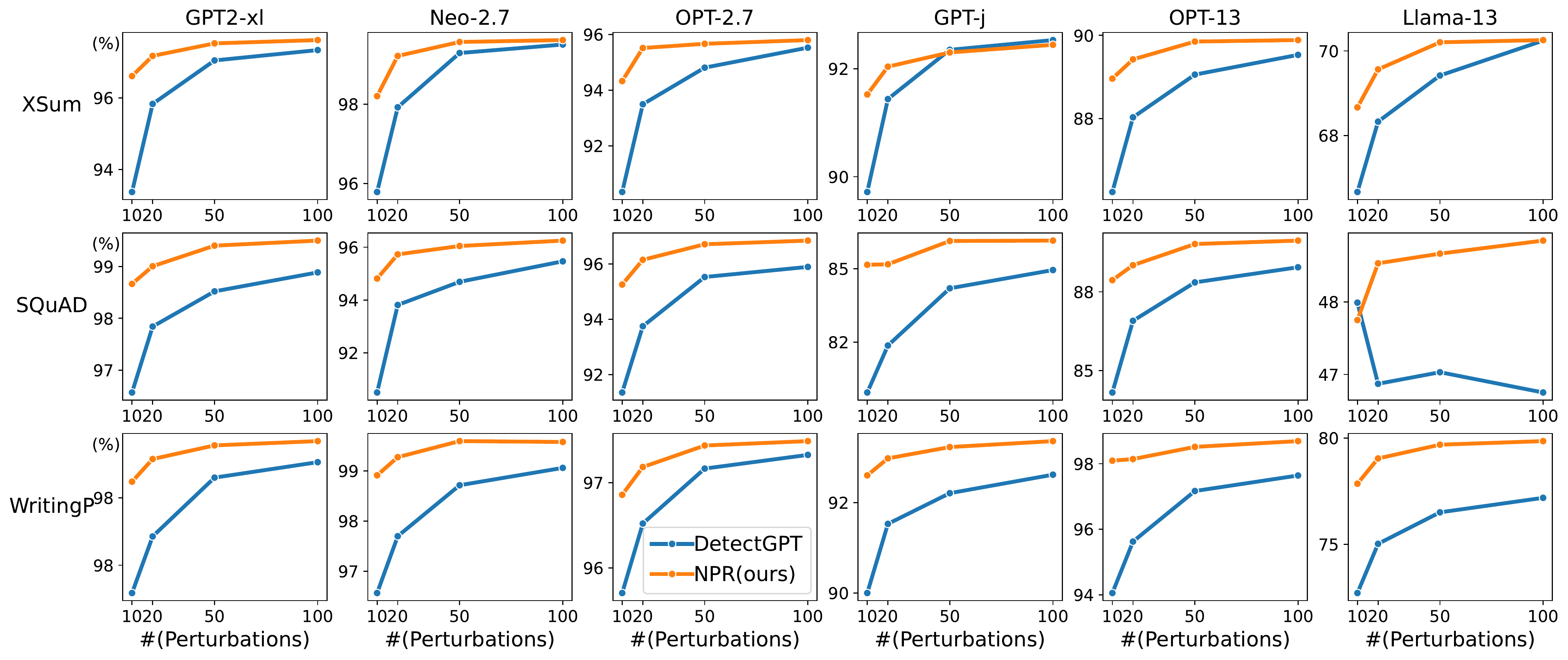}
    \caption{Comparing DetectGPT and \methodtwo (AUROC score).}
    \label{fig: rank vs prob- 6 models}
\end{figure*}

\begin{figure}[t!]
    \centering
    \includegraphics[width=0.5\textwidth]{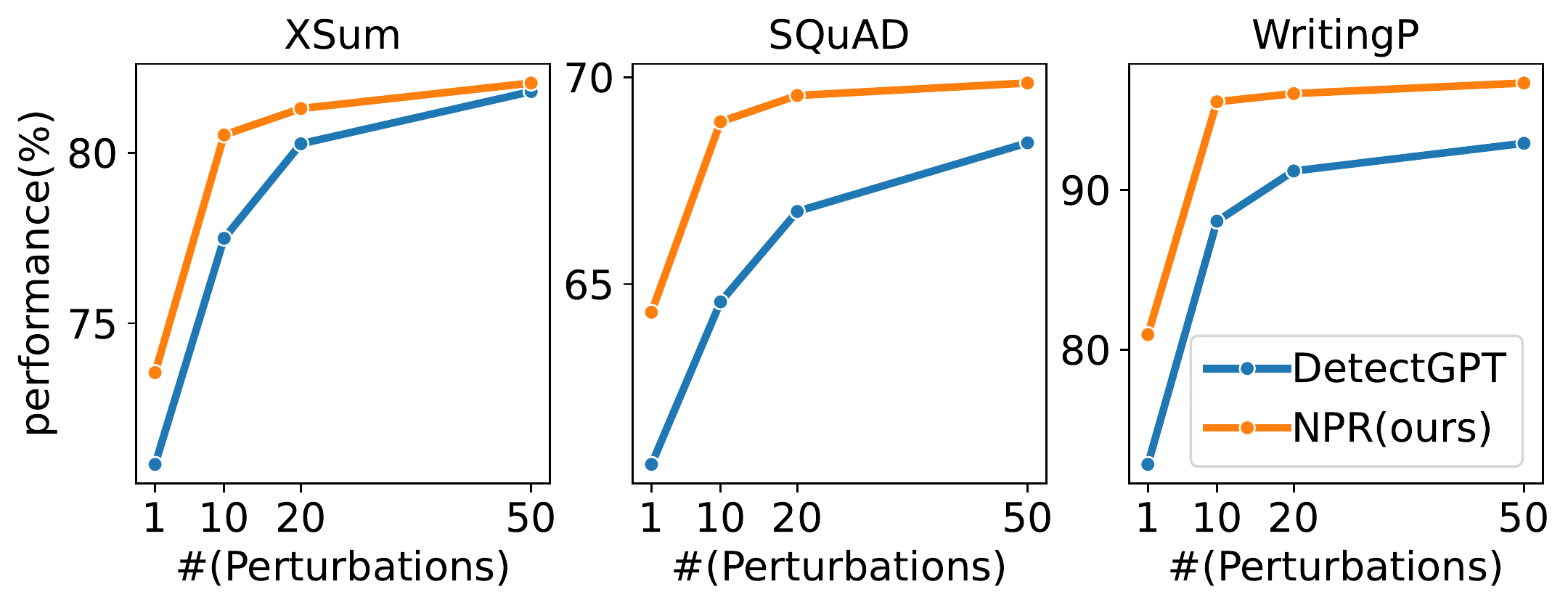}
    \caption{Comparing DetectGPT and \methodtwo on NeoX-20b (AUROC score).}
    \label{fig: rank vs prob- neoX-20b}
\end{figure}

\paragraph{Different Number of Perturbations.} The results for models smaller than or equal to 13B parameters are shown in Figure \ref{fig: rank vs prob- 6 models}. For NeoX-20b model, we don't have enough computation resources to perform 100 perturbations, so we show it separately in Figure \ref{fig: rank vs prob- neoX-20b} with 1, 10, 20, and 50 perturbations. For XSum dataset, \methodtwo and DetectGPT almost coverages with 100 perturbations, but for SQuAD and WritingPrompts dataset, \methodtwo still outperforms DetectGPT even with 100 perturbations. 
For SQuAD dataset with Llama-13b model, DetectGPT exhibits abnormality while \methodtwo maintains stably improved performance as the number of perturbations increases. 
In addition, in nearly all the dataset and models, \methodtwo outperforms DetectGPT except GPT-j on XSum dataset, demonstrating the effectiveness of \methodtwo compared to DetectGPT.

\begin{figure*}[t]
    \centering
    \includegraphics[width=1\textwidth]{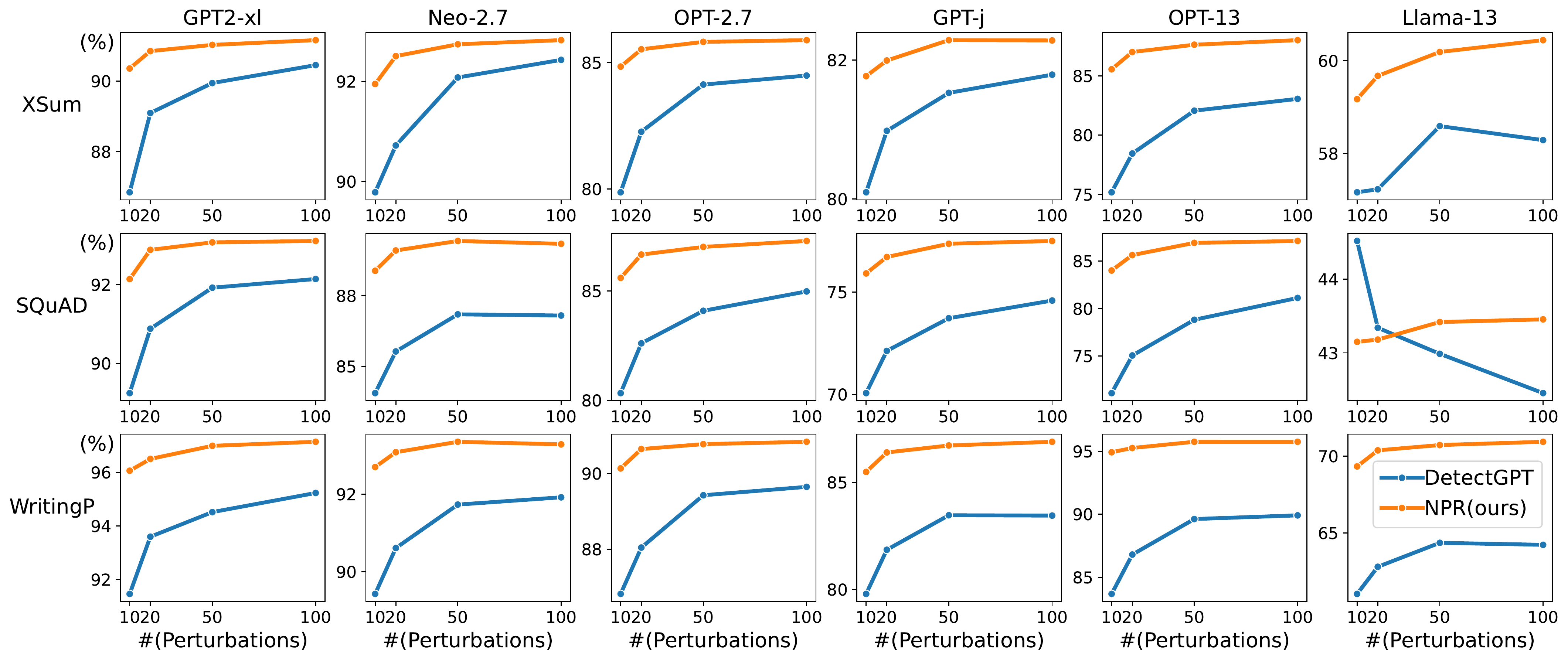}
    \caption{Comparing DetectGPT and \methodtwo using t5-large (AUROC score).}
    \label{fig: rank vs prob- t5-large}
\end{figure*}

\paragraph{Using T5-large as Perturbation Function.}
We illustrate the performance of \methodtwo and DetectGPT in Figure \ref{fig: rank vs prob- t5-large} with different combination of dataset and LLMs using T5-large as perturbation function. Compared to T5-3b illustrated in Figure \ref{fig: rank vs prob- 6 models}, the superiority of \methodtwo over DetectGPT becomes more distinct with T5-large being the perturbation function, where in almost all the LLMs, datasets and different number of perturbations (except with Llama-13b on SQuAD), \methodtwo outperforms DetectGPT by a large margin. In addition, we could also observe that \methodtwo achieves comparable or even better result with only 10 perturbations to that of DetectGPT with 100 perturbations, which indicates that \methodtwo is more efficient and can achieve similar level of performance with significantly fewer number of perturbations.

\section{Alternative Sampling Strategies and Temperature}\label{Appe: sampling strategy and temperature}
\paragraph{Different Sampling Strategy.} In Table \ref{tab: top-k, top-p}, we illustrate the complete results with different zero-shot methods with four LLMs using top-$p$ and top-$k$ sampling. For perturbation based methods, even with different sampling strategy, \methodtwo provides clearer signal for machine generated text detection than DetectGPT. Moreover, we find that although \methodone is more stable than Log Rank and Log Likelihood methods: when replacing temperature sampling to top-$p$ and top-$k$ sampling, all the above-mentioned three zero-shot methods' performance improve, however, \methodone improves approximately the same for both top-$k$ and top-$p$ sampling while the other two is more in favor of top-$p$ sampling. 
\begin{table*}
\centering
\setlength{\tabcolsep}{2pt} 
\begin{tabular}{lp{1.8cm}p{1.8cm}p{1.35cm}p{1.35cm}p{1.1cm}p{1.5cm}|p{1.35cm}p{1.35cm}p{1.1cm}p{1.5cm}}
\toprule
&&&\multicolumn{4}{c}{top-$k$}&\multicolumn{4}{c}{top-$p$}\\
\hline
Dataset&Perturbation& Method& Neo-2.7&OPT-2.7 & GPT-j&Llama-13&Neo-2.7&OPT-2.7 & GPT-j&Llama-13\\
\hline
\multirow{7}{*}{XSum}&\multirow{5}{*}{w/o}&$\log p$ &
91.27&
90.19&
85.95&
59.14&
95.52&
93.27&
91.13&
67.86 \\

&&Rank&
78.79&
76.75&
77.25&
49.94&
78.58&
76.89&
77.18&
50.77 \\

&&Log Rank&
94.20&
92.30&
89.18&
65.09&
\textbf{96.71}&
\textbf{93.93}&
\textbf{92.53}&
71.44 \\

&&Entropy&
53.07&
47.80&
53.23&
67.76&
49.05&
46.41&
52.16&
67.94 \\

&&\methodone(ours)&
\textbf{95.50}&
\textbf{92.35}&
\textbf{91.14}&
\textbf{77.99}&
95.64&
90.68&
91.14&
\textbf{75.72} \\

\cline{2-11}&\multirow{2}{*}{w/}&DetectGPT&
98.94&
96.63&
\textbf{96.56}&
73.22&
98.82&
97.72&
96.58&
77.82 \\

&&\methodtwo(ours)&
\textbf{99.61}&
\textbf{98.23}&
96.41&
\textbf{77.48}&
\textbf{99.27}&
\textbf{98.40}&
\textbf{97.35}&
\textbf{78.67} \\ 
\hline\hline
\multirow{7}{*}{SQuAD}&\multirow{5}{*}{w/o}&$\log p$ &
87.85&
91.00&
81.32&
45.06&
91.20&
94.24&
86.69&
56.16 \\

&&Rank&
80.10&
82.14&
79.81&
55.21&
80.56&
82.40&
80.28&
56.89 \\

&&Log Rank&
92.58&
94.40&
86.94&
51.21&
94.48&
96.37&
90.44&
60.66 \\

&&Entropy&
54.62&
50.83&
56.89&
69.52&
54.51&
50.01&
55.67&
63.26 \\

&&\methodone(ours)&
\textbf{97.79}&
\textbf{97.58}&
\textbf{94.55}&
\textbf{72.52}&
\textbf{97.48}&
\textbf{98.11}&
\textbf{94.38}&
\textbf{74.38} \\

\cline{2-11}&\multirow{2}{*}{w/}&DetectGPT&
97.04&
97.53&
87.59&
47.52&
97.50&
97.48&
88.90&
52.06 \\

&&\methodtwo(ours)&
\textbf{98.56}&
\textbf{99.35}&
\textbf{91.21}&
\textbf{50.83}&
\textbf{98.32}&
\textbf{99.18}&
\textbf{92.99}&
\textbf{54.28} \\ 

\hline\hline

\multirow{7}{*}{WritingP}&\multirow{5}{*}{w/o}&$\log p$ &
96.62&
95.99&
95.67&
86.93&
98.16&
98.10&
97.11&
92.68 \\

&&Rank&
82.67&
83.96&
83.49&
78.49&
82.89&
84.45&
83.55&
79.01 \\

&&Log Rank&
97.90&
97.23&
97.20&
90.57&
\textbf{98.73}&
\textbf{98.60}&
\textbf{97.89}&
94.56 \\

&&Entropy&
32.37&
38.22&
34.37&
44.09&
27.08&
36.77&
32.82&
39.03 \\

&&\methodone(ours)&
\textbf{98.58}&
\textbf{97.97}&
\textbf{98.06}&
\textbf{93.80}&
98.46&
97.97&
97.76&
\textbf{94.79} \\

\cline{2-11}&\multirow{2}{*}{w/}&DetectGPT&
99.05&
98.65&
96.05&
81.83&
98.80&
98.62&
96.67&
82.70 \\

&&\methodtwo(ours)&
\textbf{99.58}&
\textbf{99.46}&
\textbf{98.27}&
\textbf{87.99}&
\textbf{99.36}&
\textbf{99.04}&
\textbf{97.85}&
\textbf{89.96} \\ 
\bottomrule
\end{tabular}
\caption{Complete result for the zero-shot methods using top-$k$ and top-$p$ sampling across four models (AUROC score).}
\label{tab: top-k, top-p}
\end{table*}

\begin{figure*}[t]
    \centering
    \includegraphics[width=1\textwidth]{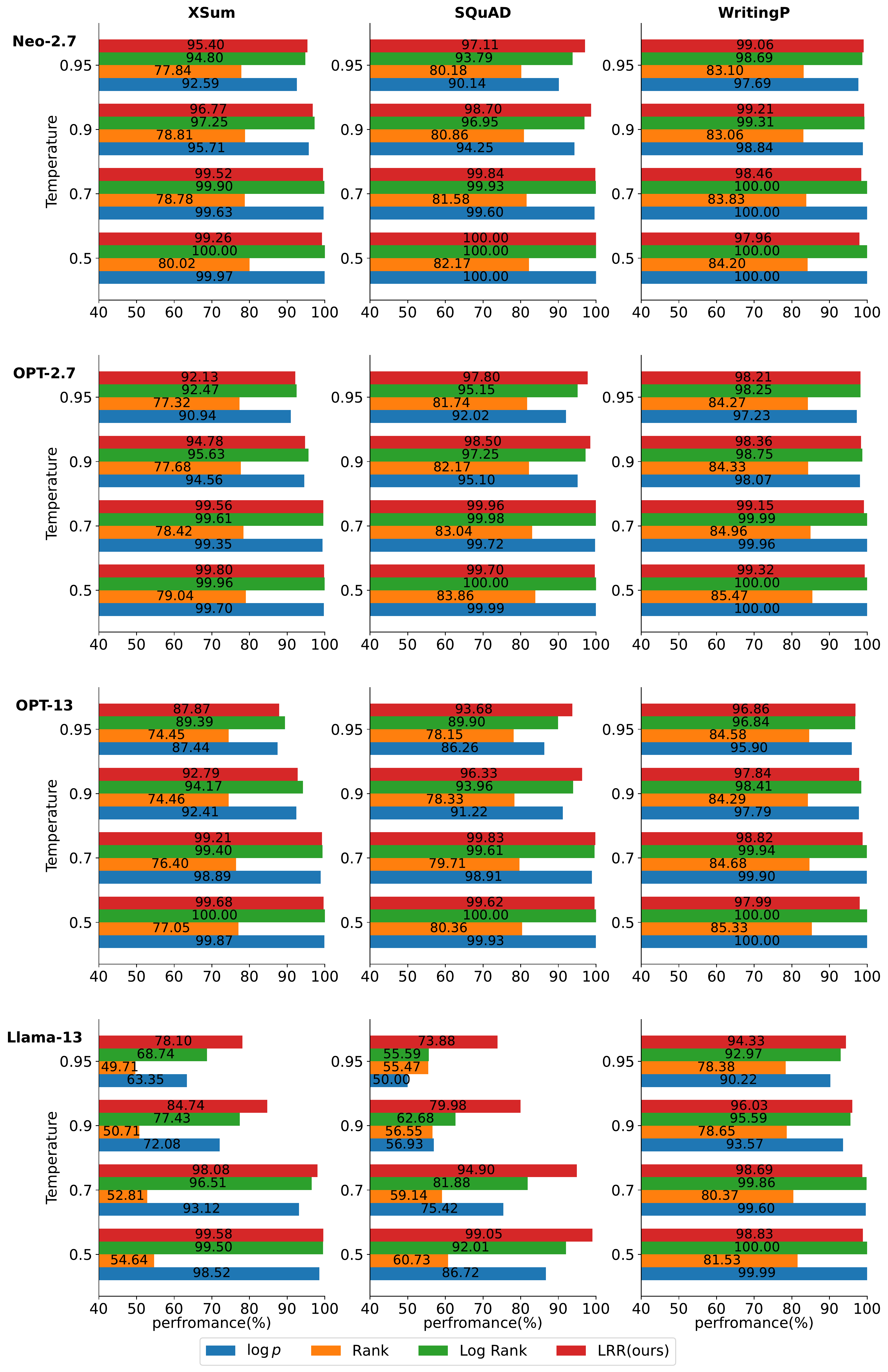}
    \caption{Comparison of perturbation-free methods using different temperature (AUROC score).}
    \label{fig: temprature-non-perturbation-based}
\end{figure*}

\paragraph{Different Temperature.} Here, we investigate how temperature used for machine generated texts affects detection accuracy of different zero-shot methods.
From Figure \ref{fig: temprature-non-perturbation-based}, we find that, all the perturbation-free zero-shot methods improved their performance with the decreasing temperature. In particular, for Log Rank and Log Likelihood method, the performance can become extremely high when the temperature drops, even exceeding \methodtwo and achieving approximately 100 points detection accuracy. For example, in Neo-2.7 and OPT-13 with temperature 0.5, $\log p$  method and Log Rank method achieves an accuracy of 100 points on WritingPrompts dataset, this prevalent performance can be observed notably in smaller models with relatively high temperature (such as GPT2-xl and Neo-2.7 with high temperature such as 0.7) or in large models with relatively lower temperature such as OPT-13 with temperature 0.5 as we demonstrated in Figure  \ref{fig: temprature-non-perturbation-based}. Though we omit entropy method because it gets an accuracy worse than random guessing, one of the observation from our experiments is that, using the assumption ``machine generated text has higher entropy" suggested in \cite{mitchell2023detectgpt}, the performance of entropy method improve with the increasing temperature with absolute accuracy smaller than 50 points, which suggests that for low temperature, we should use the assumption ``machine generated text has lower entropy" for detection machine generated text. In general, Entropy method performs worse than random and is not an implementable detection method.

\begin{figure*}[t]
    \centering
    \includegraphics[width=1\textwidth]{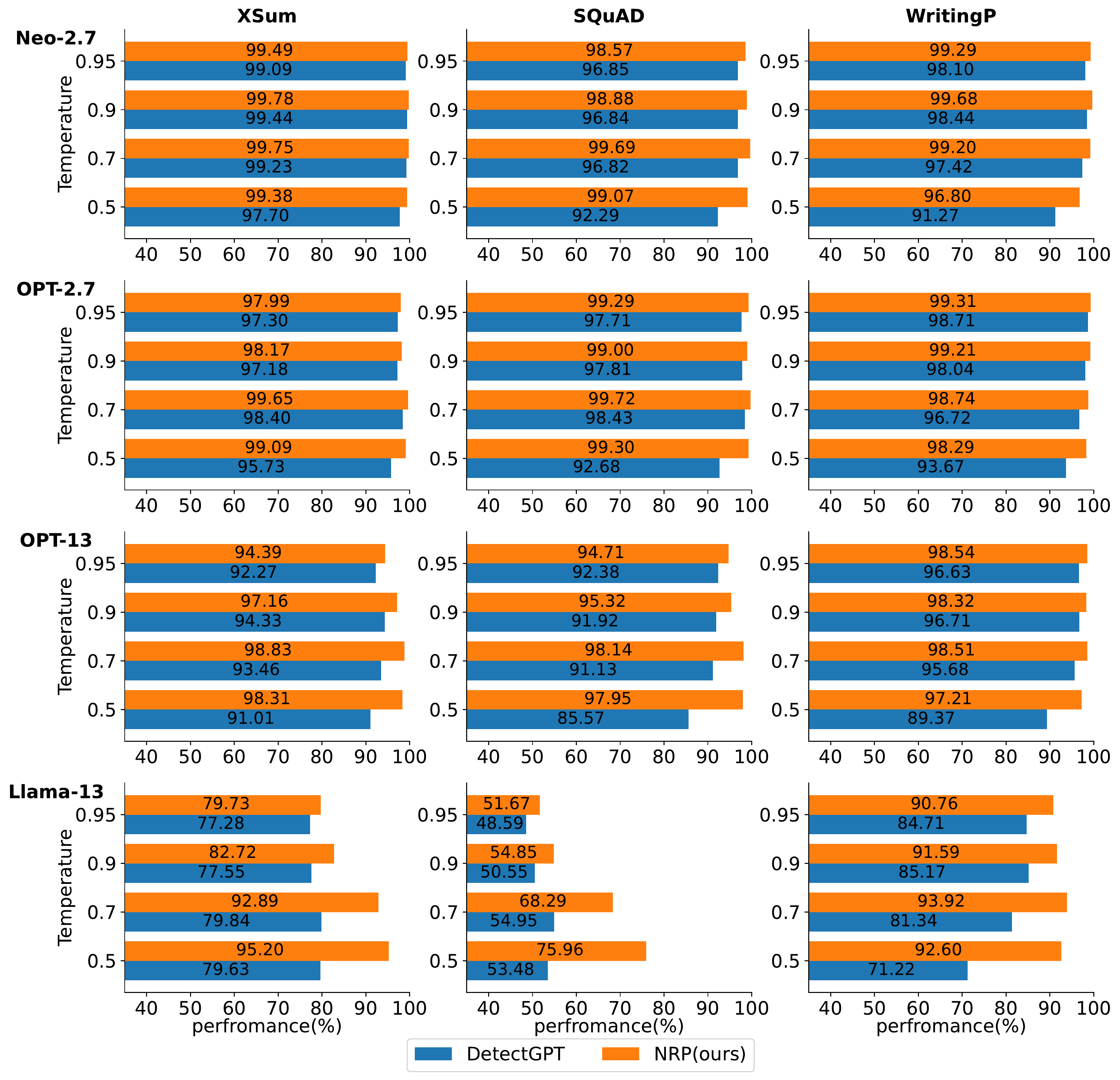}
    \caption{Comparison of perturbation methods using different temperature (AUROC score).}
    \label{fig: temprature-perturbation-based}
\end{figure*}

For perturbation based methods (Figure \ref{fig: temprature-perturbation-based}), while DetectGPT does not exhibit a clear trend with respect to temperature, the performance of \methodtwo improves with the decreasing temperature most of the time. However, this trend is not as clearly as Log Rank and log likelihood method, especially when the temperature becomes too low. This behavior suggests that perturbation based method is more suitable for high temperature, while perturbation-free method is more suitable for low temperature.

\section{Computational Time}\label{appe: computational time}
The estimated computational time of different zero-shot methods is illustrated in Table \ref{tab:computational time}. The time is estimated over the average of 10 samples. For perturbation based methods, since the time depends on the perturbation function and the number of perturbation, we used T5-3b as perturbation function and use 50 perturbations since this is the setting used for the main results in Table \ref{tab:main}, we want to provide an idea of how much more it costs for perturbation based method to achieve exceptional performance in Table \ref{tab:main}.

\begin{table*}[t!]
\centering
\setlength{\tabcolsep}{2.5pt} 
\begin{tabular}{p{1.8cm}p{1.8cm}p{1.4cm}p{1.4cm}p{1.4cm}p{1.4cm}p{1.4cm}p{1.5cm}p{1.4cm}}
\toprule
Perturbation& Method& GPT2-xl& Neo-2.7& OPT-2.7&GPT-j&OPT-13&Llama-13& NeoX\\
\midrule
\multirow{5}{*}{w/o}&$\log p$ &
0.06
&
0.09
&
0.10
&
0.04
&
0.07
&
0.07
&
0.60
\\
&Rank&
0.07
&
0.10
&
0.09
&
0.04
&
0.05
&
0.07
&
0.60
\\
&Log Rank&
0.06
&
0.09
&
0.10
&
0.04
&
0.05
&
0.06
&
0.60
\\
&Entropy&
0.06
&
0.09
&
0.09
&
0.04
&
0.05
&
0.06
&
0.60
\\
&\methodone(ours)&
0.12
&
0.19
&
0.18
&
0.08
&
0.10
&
0.14
&
1.20
\\
\midrule\multirow{2}{*}{w/}&DetectGPT&
8.07
&
9.60
&
9.80
&
7.03
&
7.98
&
8.14
&
35.56
\\
&\methodtwo(ours)&
8.15
&
9.69
&
9.90
&
7.12
&
7.83
&
7.98
&
35.67
\\\bottomrule

\end{tabular}
\caption{Computational time (seconds) for different zero-shot methods on different LLMs (averaged over 10 reruns).}
\label{tab:computational time}
\end{table*}

\end{document}